%% file: main_icra.tex
\documentclass[letterpaper, 10 pt, conference]{ieeeconf}  % Comment this line out if you need a4paper

\IEEEoverridecommandlockouts                              % This command is only needed if 
\usepackage[utf8]{inputenc} % allow utf-8 input
\usepackage[T1]{fontenc}    % use 8-bit T1 fonts
\usepackage{hyperref}       % hyperlinks
\usepackage{url}            % simple URL typesetting
\usepackage{booktabs}       % professional-quality tables
\usepackage{amsfonts}       % blackboard math symbols
\usepackage{nicefrac}       % compact symbols for 1/2, etc.
\usepackage{microtype}      % microtypography
\usepackage{xcolor}         % colors
\usepackage{svg}
\usepackage{float}
\usepackage{graphicx}
\usepackage{subfig}
\usepackage{makecell}
\usepackage{multirow}
\usepackage{amsthm}
\usepackage{amsmath,amssymb}
\usepackage{wrapfig}
\usepackage{lipsum}

\title{Selectively Dilated Convolution for Accuracy-Preserving \\ Sparse Pillar-based Embedded 3D Object Detection}
\author{Seongmin Park$^{1}$, Minjae Lee$^{1}$, Jun Won Choi$^{2}$ and Jungwook Choi$^{1}$% <-this % stops a space
\thanks{$^{1}$Seongmin Park, Minjae Lee and Jungwook choi are with the Department of Electronic Engineering, Hanyang University, Seoul 04763, South Korea {\tt\small \{skstjdals,lmj4666,choij\}@hanyang.ac.kr}}%     
\thanks{$^{2}$JunWon Choi is with the Department of Electrical Engineering, Hanyang University, Seoul 04763, South Korea {\tt\small junwchoi@hanyang.ac.kr}}%
}

\begin{document}

\maketitle
\thispagestyle{empty}
\pagestyle{empty}

%%%%%%%%%%%%%%%%%%%%%%%%%%%%%%%%%%%%%%%%%%%%%%%%%%%%%%%%%%%%%%%%%%%%%%%%%%%%%%%%
\begin{abstract}

Pillar-based 3D object detection has gained traction in self-driving technology due to its speed and accuracy facilitated by the artificial densification of pillars for GPU-friendly processing. However, dense pillar processing fundamentally wastes computation since it ignores the inherent sparsity of pillars derived from scattered point cloud data. Motivated by recent embedded accelerators with native sparsity support, sparse pillar convolution methods like submanifold convolution (SubM-Conv) aimed to reduce these redundant computations by directly applying convolution on sparse pillars but suffered considerable accuracy loss. Our research identifies that this accuracy loss is due to the restricted spatial information flow of SubM-Conv in sparse pillar networks. To overcome this restriction, we propose a \textit{selectively dilated} convolution that evaluates the importance of encoded pillars and selectively expands its output for sparse convolution, enhancing the receptive field for critical pillars and improving object detection accuracy. Evaluations with existing embedded sparse convolution accelerators on various cutting-edge pillar-based 3D object detection models confirm that our approach achieves extreme pillar sparsity, leading to up to 18.1$\times$ computational savings and 16.2$\times$ speedup on the embedded accelerators, without compromise in object detection accuracy.

\end{abstract}

%%%%%%%%%%%%%%%%%%%%%%%%%%%%%%%%%%%%%%%%%%%%%%%%%%%%%%%%%%%%%%%%%%%%%%%%%%%%%%%%
\input{Sections/introduction}
\input{Sections/related}
\input{Sections/background}

\input{Sections/method}
\input{Sections/experiments}
\input{Sections/conclusion}

\addtolength{\textheight}{-8.3cm}   % This command serves to balance the column lengths
                                  % on the last page of the document manually. It shortens
                                  % the textheight of the last page by a suitable amount.
                                  % This command does not take effect until the next page
                                  % so it should come on the page before the last. Make
                                  % sure that you do not shorten the textheight too much.

%%%%%%%%%%%%%%%%%%%%%%%%%%%%%%%%%%%%%%%%%%%%%%%%%%%%%%%%%%%%%%%%%%%%%%%%%%%%%%%%

%%%%%%%%%%%%%%%%%%%%%%%%%%%%%%%%%%%%%%%%%%%%%%%%%%%%%%%%%%%%%%%%%%%%%%%%%%%%%%%%

%%%%%%%%%%%%%%%%%%%%%%%%%%%%%%%%%%%%%%%%%%%%%%%%%%%%%%%%%%%%%%%%%%%%%%%%%%%%%%%%
% \section*{APPENDIX}

% Appendixes should appear before the acknowledgment.

% \section*{ACKNOWLEDGMENT}

% The preferred spelling of the word ÒacknowledgmentÓ in America is without an ÒeÓ after the ÒgÓ. Avoid the stilted expression, ÒOne of us (R. B. G.) thanks . . .Ó  Instead, try ÒR. B. G. thanksÓ. Put sponsor acknowledgments in the unnumbered footnote on the first page.

%%%%%%%%%%%%%%%%%%%%%%%%%%%%%%%%%%%%%%%%%%%%%%%%%%%%%%%%%%%%%%%%%%%%%%%%%%%%%%%%

% References are important to the reader; therefore, each citation must be complete and correct. If at all possible, references should be commonly available publications.
\bibliographystyle{plain}
\bibliography{refs}

\end{document}

\appendix

\section{Appendix}

\subsection{Important Pillar Selection Based on $I_p$ }
In section 3.1, when choosing which pillar to perform dilation on based on pillar importance ($I_p$), we use the top-k method. As can be seen in Table~\ref{tab:t_sweep}, choosing $t=2$ in PointPillars provides a good trade-off between computation cost and accuracy. We observe that computational cost diminishes as move from $t=5$ to $t=1$. The mAP remains relatively constant when $t$ is reduced from 5 to 2. However, there's a steep decline in mAP at $t=1$, suggesting a significant compromise in accuracy despite the lowered computational cost.
\begin{table}[h]
\caption{Performance comparison for the $t\%$ in top-k selection on KITTI \textit{val} set with PointPillars.}
\begin{center}
\resizebox{0.6 \linewidth}{!}{
\renewcommand{\arraystretch}{1}
\begin{tabular}{c|cccccc}
\Xhline{2\arrayrulewidth}
\textit{t}  & \multicolumn{1}{l}{baseline}     & 5    & 4     & 3     & 2     & 1     \\ \hline
2D Hard mAP & 85.63                  & \textbf{85.80} & 85.75 & 85.66 & 85.61 & 83.38 \\ \hline
FLOPs(G)    & 46.43                  & 7.72 & 7.39  & 6.99  & 6.43  & \textbf{5.91}  \\ \Xhline{2\arrayrulewidth}
\end{tabular}}
\end{center}
\label{tab:t_sweep}
\vspace{-0.5 cm}
\end{table}

\subsection{Detailed Model Structure}
\begin{figure}[H]
\centering
\includegraphics[width=0.85\linewidth]{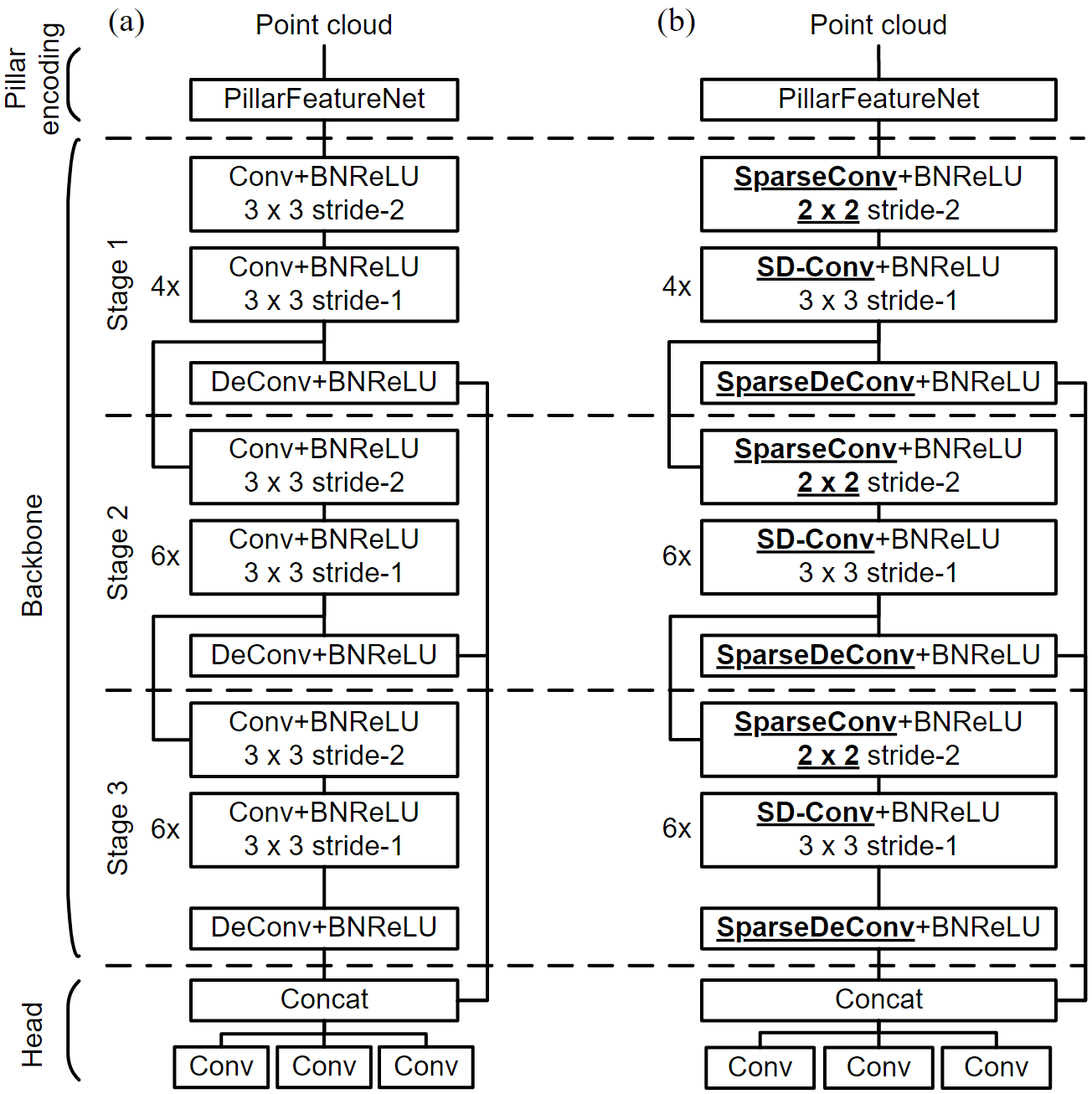}
%\includesvg[inkscapelatex=false,width=0.71\linewidth]{Figure/appen_fig1_2.svg}
\caption{(a) Basic structure of PointPillars, (b) Applying SD-Conv in PointPillars.}
\label{fig_struc_PP}
\end{figure}

For the convenience of readers, we provide the detailed architecture of our models. We compare two models: the basic one and another with SD-Conv applied. In the case of PointPillars, as shown in Fig.~\ref{fig_struc_PP}, all the convolution modules in the backbone, which include the stride-2 Conv with a 3x3 kernel size, the stride-1 Conv, and the DeConv, are replaced with a 2x2 stride-2 SparseConv, SD-Conv, and SparseDeConv, respectively.

\begin{figure}[t]
\centering
\includegraphics[width=0.85\linewidth]{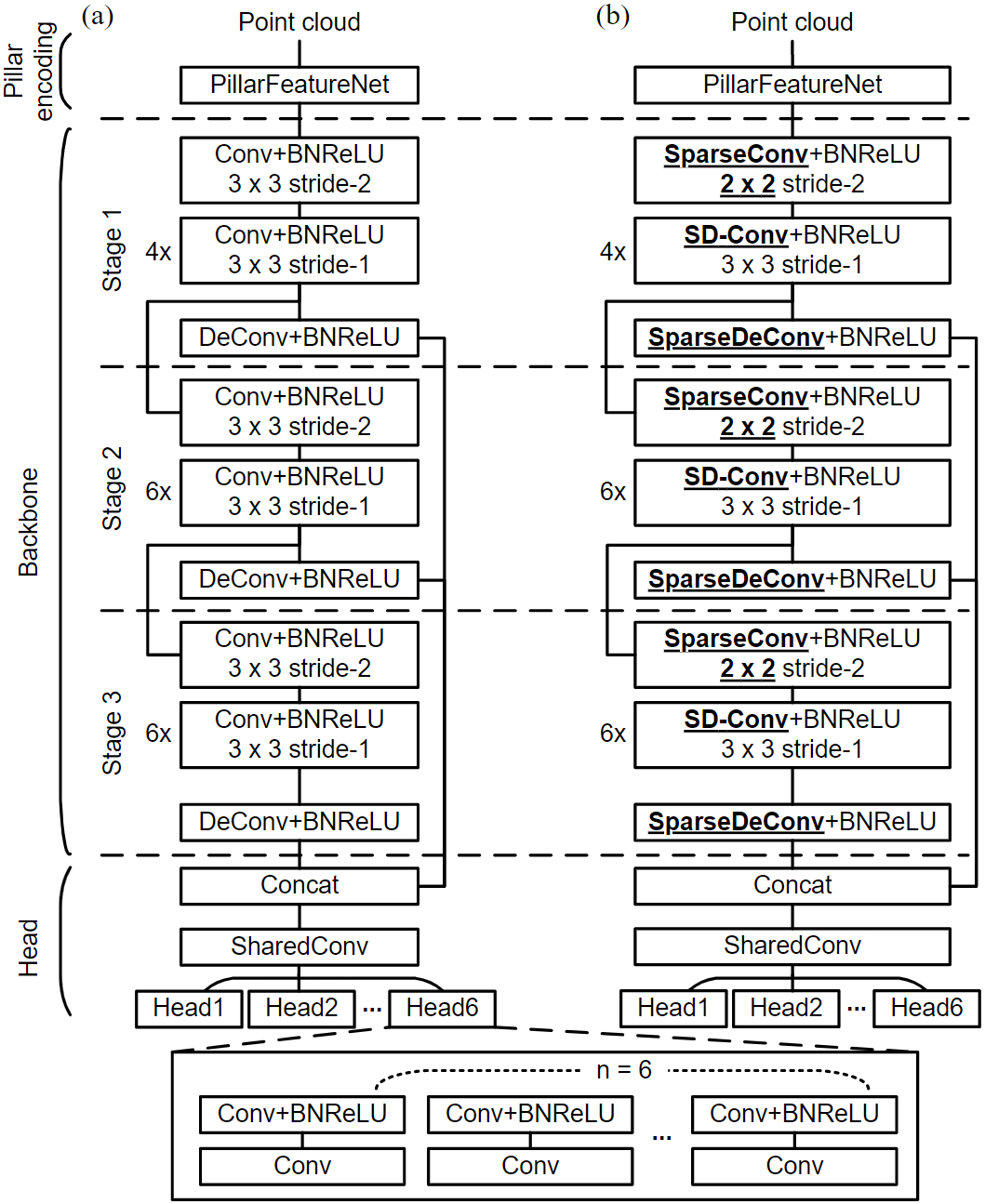}
%\includesvg[inkscapelatex=false,width=0.77\linewidth]{Figure/appen_fig2_2.svg}
\caption{(a) Basic structure of CenterPoint, (b) Applying SD-Conv in CenterPoint.}
\label{fig_struc_CP}
\end{figure}

For CenterPoint, as depicted in Fig.~\ref{fig_struc_CP}, it has the same backbone structure as PointPillars.  However, its head structure differs as it adopts a multi-head architecture with individual heads assigned to each object type. In the model with SD-Conv applied, the CenterPoint backbone's convolution operations are replaced with sparse convolution modules that include SD-Conv, similar to the approach in PointPillars.

\begin{figure}[H]
\centering
\includegraphics[width=\linewidth]{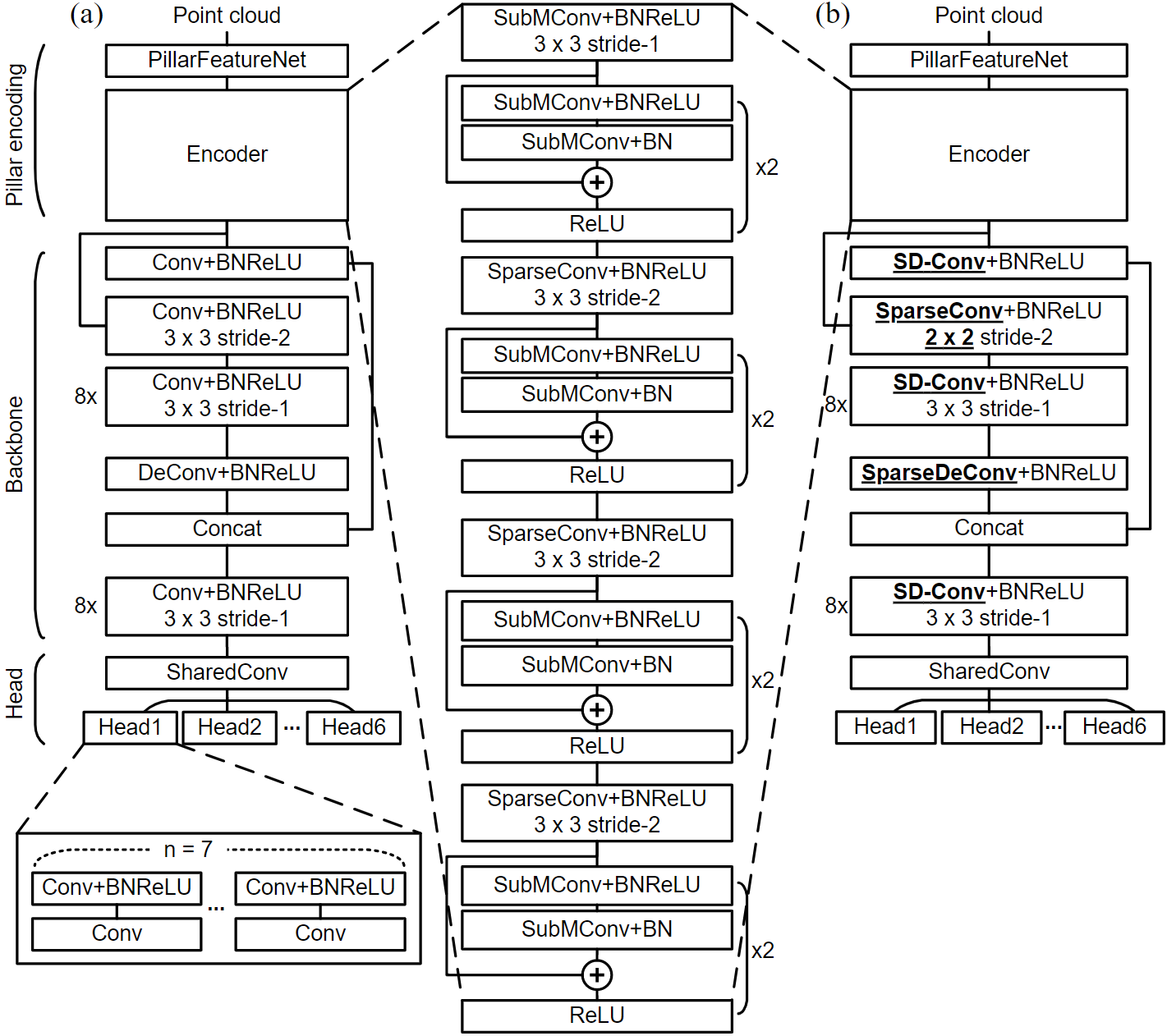}
%\includesvg[inkscapelatex=false,width=\linewidth]{Figure/appen_fig3_2.svg}
\caption{(a) Basic structure of PillarNet, (b) Applying SD-Conv in PillarNet.}
\label{fig_struc_PN}
\end{figure}

As shown in Fig.~\ref{fig_struc_PN}, PillarNet has an advanced encoder structure for deep pillar feature extraction. Its backbone structure also diverges from PointPillars and CenterPoint. However, similar to CenterPoint, PillarNet's head structure utilizes a multi-head approach. In the model where SD-Conv is employed, we replace Conv modules of backbone with a sparse convolution module that includes SD-Conv. This introduction of SD-Conv effectively reduces the backbone density by 27.05\%, and brings down the overall FLOPs from 283.98G to 155.02G.
% \begin{figure}[h]
% \centering
% \includegraphics[width=0.85\linewidth]{Figure/appen_fig1.png}
% %\includesvg[inkscapelatex=false,width=0.71\linewidth]{Figure/appen_fig1_2.svg}
% \caption{(a) Basic structure of PointPillars, (b) Applying SD-Conv in PointPillars.}
% \label{fig_struc_PP}
% \end{figure}

\subsection{Applying Pruning from Scratch in Training: SDP-Conv }
%Another consideration related to SPP is that pruning can be applied from scratch during the training process, which we denote as SDP-Conv. Training with pruning applied from the scratch makes optimal learning challenging, as pillars are removed before the backbone network's feature extraction capabilities become powerful, leading to performance degradation, as seen in Table~\ref{tab:sdp}.
In terms of pruning, another consideration is the possibility of applying pruning from the beginning of the training process, a method we denote as SDP-Conv. The SDP-Conv module, utilizing both $I_p^{S\!P\!P}$ and $I_p^{S\!D}$, determines which pillars should be pruned and which should be subjected to dilate. This module is more general as it takes into account both pruning and dilation simultaneously. However, training with pruning applied from the scratch poses a challenge for achieving optimal learning. This is due to the removal of pillars before the feature extraction capabilities of the backbone network become sufficiently powerful, which results in a degradation in performance, as can be seen in Table~\ref{tab:sdp}.

\begin{table}[H]
\caption{SD-Conv + SPP vs SDP on KITTI \textit{val} set with PointPillars.}
\begin{center}
\resizebox{0.7 \linewidth}{!}{
\renewcommand{\arraystretch}{1.1}
\begin{tabular}{c|c|ccc|ccc}
\Xhline{2\arrayrulewidth}
\multirow{2}{*}{Method}       & \multirow{2}{*}{\begin{tabular}[c]{@{}c@{}}FLOPs\\ (G)\end{tabular}} & \multicolumn{3}{c|}{BEV} & \multicolumn{3}{c}{3D} \\ \cline{3-8} 
                              &                                                                      & Easy   & Mod    & Hard   & Easy   & Mod   & Hard  \\ \Xhline{2\arrayrulewidth}
SD-Conv + SPP                       & \textbf{4.46}                                                                 & \textbf{90.29}  & \textbf{87.63}  &\textbf{85.45}  & \textbf{87.44}  & \textbf{77.22} & \textbf{75.08} \\ \hline
\multicolumn{1}{c|}{SDP-Conv} & 4.64                                                                 & 90.00     & 87.34  & 84.45  & 86.88  & 76.99 & 74.59 \\ \Xhline{2\arrayrulewidth}
\end{tabular}}
\end{center}
\label{tab:sdp}
\vspace{-0.5 cm}
\end{table}

\clearpage

% \bibliographystyle{unsrt}
% \bibliography{reference}
% \iffalse
%     \section*{References}
%     \small
% \fi

%% file: Sections/introduction.tex
% \begin{figure*}[t]
% \centering
% % \includegraphics[width=\linewidth]{Figure/dialte_draft.png}
% % \includegraphics[width=\linewidth]{Figure/dialte_draft2.png}
% \includegraphics[width=\linewidth]{Figure/dialte_draft4.png}
% \caption{(a) Pillar-based 3D object detection network structure. (b) Feature extraction steps: Backbone, Neck, and Head. (c) Comparison of the receptive field of various sparse convolution operations within a stage: Dense-Conv, SubM/SPS-Conv, FS-Conv, and SD-Conv.}
% \label{fig:fig_pillar-obj-det}
% \end{figure*}

\section{Introduction}
With the shift of priorities in autonomous driving from convenience to safety, there is a growing need for robust perception systems that can accurately interpret time-critical semantic information in real-time, such as identifying and locating road users~\cite{arnold2019survey}. At the heart of these systems is \textit{3D object detection} that leverages LiDAR-generated point cloud data, providing comprehensive depth information and obstacle detection~\cite{arnold2019survey, mao20223d}. In pursuit of real-time 3D object detection, research has gravitated towards grid-based methods that convert point clouds into 3D voxels or 2D pillars. One prime example is PointPillars~\cite{lang2019pointpillars}, which utilizes a bird's-eye-view encoding technique, aggregating 3D point cloud features into sparse 2D pillars and transforming them into a dense pseudo-image for GPU-friendly 2D convolution (Conv2D). Thanks to this improved GPU efficiency, PointPillars has emerged as a leading solution for time-critical 3D object detection~\cite{zhou2020end, sun2020scalability, shi2022pillarnet, wang2020pillar,   li2021lidar,yin2021center}. 

However, emphasizing the fundamental inefficiency in dense pillar processing, which disregards the inherent sparsity of pillars stemming from dispersed point cloud data, the emergence of dedicated embedded accelerators with native support for sparse point cloud data imposes significant opportunities for sparse pillar-based object detection toward further speedup~\cite{feng2020mesorasi,lin2021pointacc,feng2022crescent,lee2024spade}. A notable attempt is SparsePointPillars~\cite{vedder2021sparse}, which introduces submanifold convolution (SubM-Conv~\cite{graham2017submanifold}) to sparsify pillar representation and significantly reduces the number of computations. Although SubM-Conv effectively preserves point cloud's original sparsity by limiting convolution dilation, the improved sparsity comes at the cost of significant degradation in 3D object detection accuracy; as shown in Fig.~\ref{fig_spar_acc2}, Sparse PointPillars, which employs SubM-Conv, results in significant accuracy loss, especially for the complex mode (Hard). Therefore, a comprehensive understanding of the trade-off between sparsity and accuracy on sparse pillar convolution is lacking.

In this work, we identify that the key cause of the accuracy loss of previous sparse pillar convolutions is the sparsification structure that limits the \textit{fine-grained spatial information flow} (f-SIF) from the increase of receptive field via dilation. Note that prior works on voxel-based methods \cite{chen2022focalconv,liu2022spatial,chen2023voxenext} have noticed a similar issue on the SIF, but they primarily have focused on the extension of the receptive field by strided sparse convolution only at each stage of 3D object detection backbone. This approach only enhances the \textit{coarse-grained} SIF, thus showing a limited improvement in accuracy. In contrast, we reveal that the increase of the receptive field via \textit{fine-grained} selective dilation at every convolution layer within the same stage plays a crucial role in constructing necessary receptive fields for identifying 3D objects in the scene. Therefore, we propose a simple yet effective operation called \textit{selectively dilated convolution} (SD-Conv) that can identify important pillars at every convolution layer, based on their magnitude for selective dilation.
%We further propose a novel pillar-pruning method called submanifold pillar pooling (SPP) that measures the necessity of an existing pillar based on its neighboring submanifolds on the fly (without fine-tuning), granting opportunities for newly dilated pillars to connect the SIF as a group while pruning unnecessary pillars. 
%The selective dilation of SD-Conv 
%By exploiting existing sparse point cloud accelerators, 
%which is our custom-developed sparse point cloud accelerator,

To achieve actual speedup, we have designed a specialized sparse point cloud accelerator architecture that operates in a streaming manner to support SD-Conv for acceleration.
We evaluate the proposed method on various state-of-the-art pillar-based 3D object detection networks, including PointPillars~\cite{lang2019pointpillars}, CenterPoint~\cite{yin2021center}, and PillarNet~\cite{shi2022pillarnet}, as well as popular benchmarks like KITTI~\cite{geiger2012kitti} and Nuscene~\cite{caesar2020nuscenes}. The experimental results consistently demonstrate that the proposed SD-Conv can simply replace SubM-Conv to recover accuracy while achieving higher sparsity. The achieved accuracies are on par with or even surpass the accuracy of the dense baseline models, while reducing the number of computations by 94.5\%, 72.3\%, 41.3\% for PointPillars, CenterPoint, and PillarNet, respectively. With in-depth ablation study, we further demonstrate that our SD-Conv achieve superior accuracy-sparsity trade-offs compared to the prior sparse convolution approaches~\cite{chen2022focalconv,liu2022spatial,chen2023voxenext}. 
Moreover, when simulated on SPADE+, the method exhibited significant sparsity-proportional speedup of 16.2$\times$, 3.1$\times$, 1.7$\times$, respectively. These findings emphasize the effectiveness of our approach and its potential to enable real-time 3D object detection, making it suitable for time-critical applications such as autonomous driving.

%% file: Sections/background.tex
\begin{figure}[]
\centering
\includegraphics[width=\linewidth]{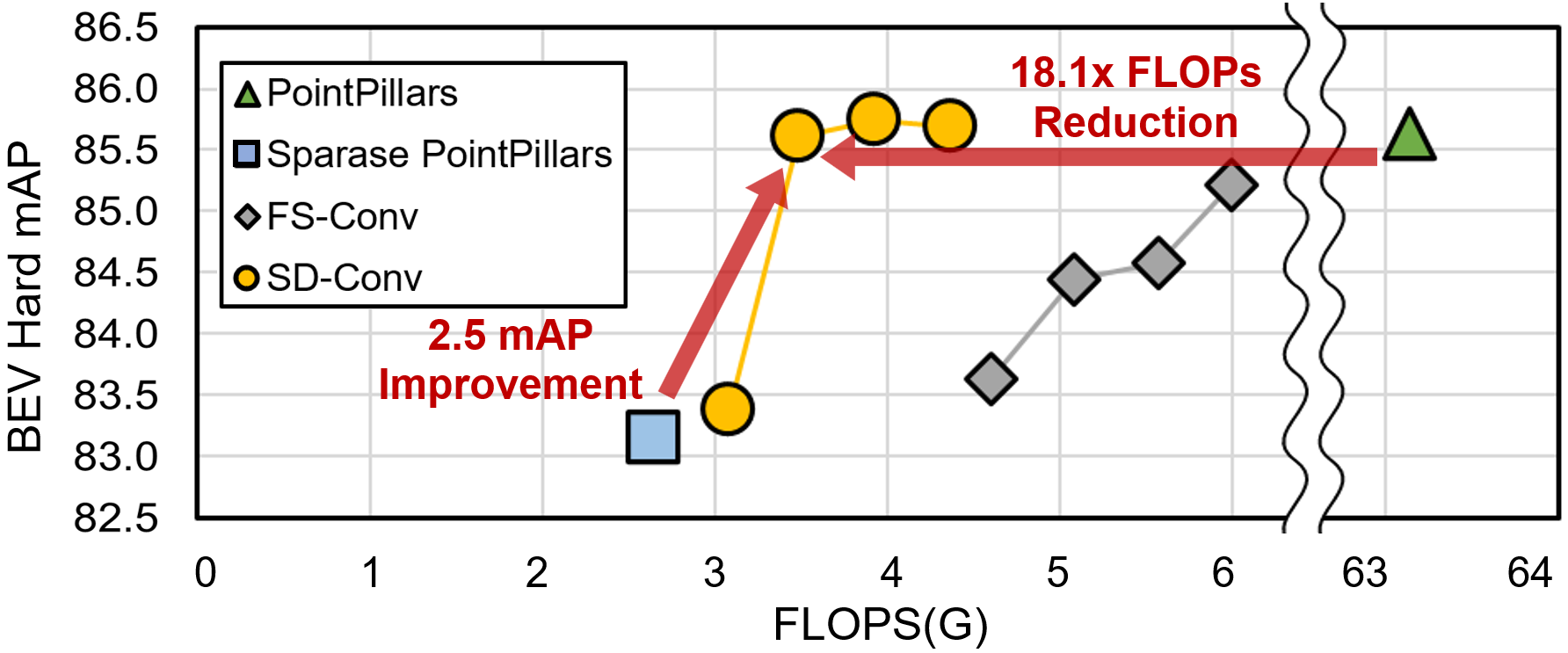}
%\includesvg[inkscapelatex=false,width=1\linewidth]{Figure/spar_acc3.svg}
% \caption{Bird's eye view performance vs FLOPs for our proposed SD-Conv and SD-Conv+SPP on KITTI \textit{val} dataset with PointPillars.}
% \caption{Bird's-eye-view performance versus FLOPs was evaluated for our proposed SD-Conv on the KITTI \textit{val} set using PointPillars.}
\caption{The accuracy and computation trade-off of 3D object detection. The pillar-based baseline, PointPillars~\cite{lang2019pointpillars}, delivers high accuracy but uses redundant computation. Sparse PointPillars~\cite{vedder2021sparse} employs SubM-Conv, reducing computation but losing considerable accuracy. In contrast to FS-Conv~\cite{chen2022focalconv}'s inferior trade-off, our selectively dilated convolution (SD-Conv) retains accuracy while cutting computations by 18.1$\times$, promising for embedded 3D object detection.}
\label{fig_spar_acc2}
\vspace{-0.1cm}
\end{figure}
\begin{figure*}[t]
\centering
\includegraphics[width=\linewidth]{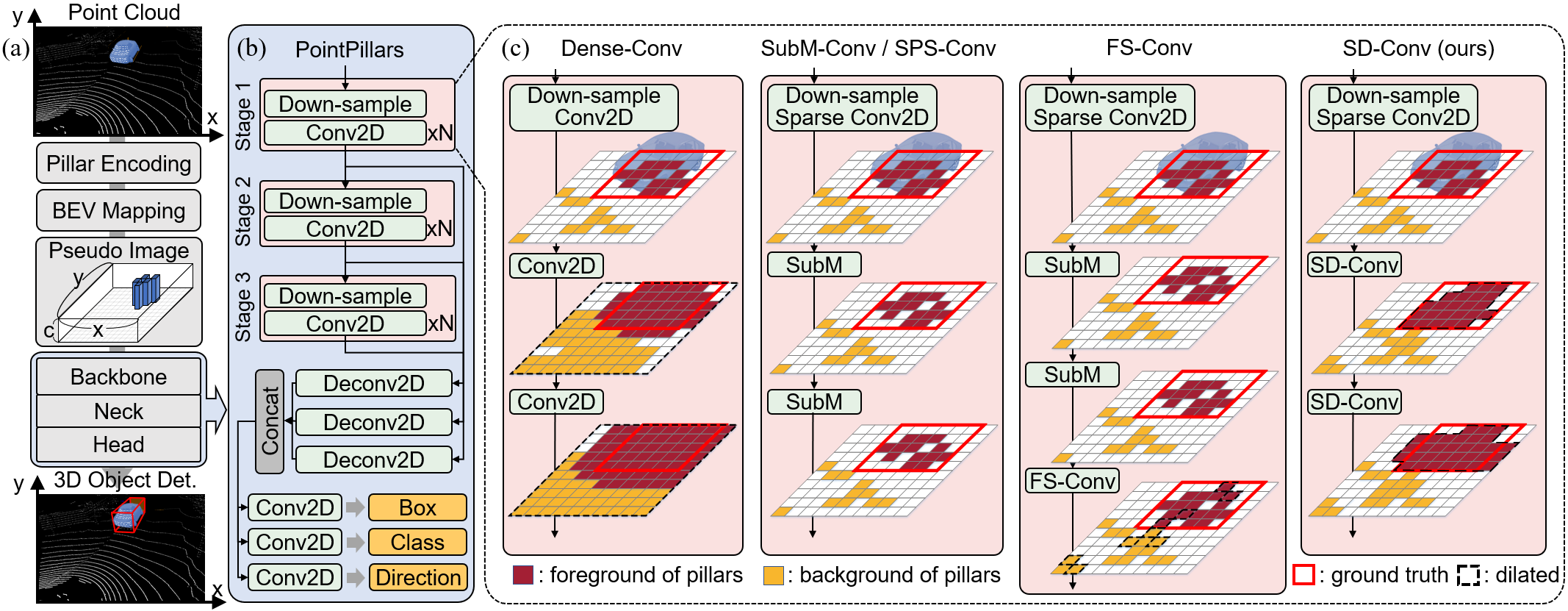}
\caption{(a) Pillar-based 3D object detection network structure. (b) Feature extraction steps: Backbone, Neck, and Head. (c) Comparison of the receptive field of various sparse convolution operations within a stage: Dense-Conv, SubM/SPS-Conv, FS-Conv, and SD-Conv.}
\label{fig:fig_pillar-obj-det}
\end{figure*}
\section{Background and Challenges}

\subsection{Pillar-based 3D Object Detection}
\label{subsec:pillar-based-3d-object-detection}

Essential for autonomous driving, 3D object detection can be implemented through a variety of approaches, including point-based, voxel-based, and pillar-based techniques. Point-based methods, such as PointNet~\cite{qi2017pointnet} and PointNet++\cite{qi2017pointnet++}, deal directly with point cloud data, but they can encounter complexity issues due to sampling and sorting processes. Conversely, voxel-based methods like VoxelNet~\cite{zhou2018voxelnet} partition space into 3D grids, but the inherent sparsity of 3D voxels can pose difficulties in GPU utilization. Pillar-based methods, such as PointPillars~\cite{lang2019pointpillars} as seen in Fig.~\ref{fig:fig_pillar-obj-det}(a), which segment space into 2D grids and utilize bird-eye-view (BEV) encoding, have risen in popularity for real-time 3D object detection~\cite{wang2020pillar, zhou2020end, sun2020scalability, li2021lidar, shi2022pillarnet,yin2021center}. However, densify sparse BEV-based 2D convolution can lead to redundant computation, suggesting that improvements can be made in PointPillars' current implementation.

% Pillar-based 3D object detection, as seen in Fig.~\ref{fig:fig_pillar-obj-det}(a), projects point clouds into a 2D space to extract features from pillars, representing bird's-eye view (BEV) features from points within a discretized region $X\times Y$, disregarding $z$ direction limits. In the pillar encoding phase, $C$-element feature vectors are derived from the point clouds via a multi-layer perceptron, leading to a 2D BEV pseudo-image of size $C\times X\times Y$. Despite the 97\% sparsity of active pillars organizing sparse points into a 2D grid, the sparse-to-dense conversion overlooks significant potential computational savings to enhance GPU efficiency for backbone and detection head computations. The PointPillars approach divides the point cloud into a 2D grid on the x-y plane to create a pillar set, which is then processed by tiny PointNets to learn features. These features are then scattered back to their original locations to generate a pseudo-image, providing an efficient representation of the point cloud data and enabling the extraction of meaningful characteristics from the pillar-based features. 

Fig.~\ref{fig:fig_pillar-obj-det}(b) illustrates the details of this feature extraction consisting of a backbone, neck, and head. The backbone consists of multiple stages of convolutions led by sparse down-sample convolution layers for the increased receptive fields. The outcome of all these stages is deconvoluted and then concatenated in the neck for box, class, and direction prediction in the head. Note that variations exist; CenterPoint~\cite{yin2021center} incorporates center-based prediction heads while PillarNet~\cite{shi2022pillarnet} strengthens pillar encoding with additional SubM-Conv layers in front. 

To achieve real-time speed, the key challenge in pillar-based methods is to reduce computations while extracting sufficient features for accurate object detection. To this end, the sparsification of pillars and utilizing sparse convolutions are getting significant attention for embedded 3D object detection since dedicated point cloud accelerators with native sparse convolution support have emerged as attractive alternatives for GPU~\cite{feng2020mesorasi,lin2021pointacc,feng2022crescent,lee2024spade}.

  % PillarNet~\cite{shi2022pillarnet} further applied SubM convolution to the additional layers of pillar encoding to strengthen the feature extraction capability within the pillar encoding stage while suppressing computation overhead. 

% \begin{figure}[]
% \centering
% \includegraphics[width=\linewidth]{Figure/spar_acc3.png}
% %\includesvg[inkscapelatex=false,width=1\linewidth]{Figure/spar_acc3.svg}
% % \caption{Bird's eye view performance vs FLOPs for our proposed SD-Conv and SD-Conv+SPP on KITTI \textit{val} dataset with PointPillars.}
% \caption{Bird's-eye-view performance versus FLOPs was evaluated for our proposed SD-Conv on the KITTI \textit{val} set using PointPillars.}
% \label{fig_spar_acc2}
% \end{figure}

\subsection{Sparse Convolution}
\label{subsec:}

Given the intrinsic sparsity of point cloud data, 3D object detection methods employ sparse 3D convolution~\cite{graham2015sparse} for efficiency. While conventional sparse convolution uses only non-zero elements in the input feature map, reducing the number of floating point operations (FLOPs) and memory demands, its dilation property can compromise overall sparsity.

\textbf{Submanifold Sparse Convolution (SubM-Conv)} ~\cite{graham2017submanifold} addresses this by forming a receptive field only around non-zero elements without dilation, further reducing computational requirements. However, this limited receptive field can lead to significant accuracy loss.

\textbf{Spatial Pruned Sparse Convolution (SPS-Conv)~\cite{liu2022spatial} \& Focal Sparse Convolution (FS-Conv)~\cite{chen2022focalconv}} both offering coarse adaptive dilation based on voxel importance only once at each stage. SPS-Conv measures importance based on magnitude, whereas FS-Conv calculates the importance of each voxel through additional parameters learned during training. Furthermore, FS-Conv also learns the importance of dilation direction using extra parameters, enabling partial dilation.

\textbf{Pruned Sparse Convolution (PS-Conv)~\cite{lee2024spade}} initially performs dilation on all non-zero values, akin to Dense-Conv as illustrated in Fig.~\ref{fig:fig_pillar-obj-det}(c), to then increase sparsity by applying pruning for computational reduction. However, employing high sparsity pruning during training can hinder stable learning, thereby imposing a limitation on accuracy.

\textbf{Hardware for Sparse Convolution: } 
To accelerate Sparse Convolution on Pillars, it's essential to operate only on non-zero values. This involves utilizing mapping information that represents the relationship between sparse input and sparse output~\cite{yan2018second}. One approach applicable to general-purpose processors like GPUs is employing hash tables. However, implementing hash tables introduces overhead from mapping that often outweighs the reduction in computation, thereby falling short of fully realizing the benefits of sparse convolution. Dedicated accelerators like PointAcc~\cite{lin2021pointacc} or SPADE~\cite{lee2024spade} address this by parallelizing the mapping process, reducing mapping overhead, and efficiently managing data to achieve speedup based on sparsity.
% In particular, SPADE proposes a novel sparse convolution, termed Pruned Sparse Convolution (PS-Conv), which further elevates sparsity through dynamic vector pruning. This method measures the importance of pillars based on their magnitude and dynamically prunes them to enhance sparsity. However, because this approach involves conducting convolution operations before applying pruning, it includes redundant computations that could limit the potential to reduce computational load. Despite achieving higher performance compared to using SubM-Conv, PS-Conv requires significantly more computational resources.
\subsection{Challenges: Spatial Information Flow (SIF) within a Stage}
\label{subsec:}

To address SparsePointPillars' limitations and the challenges of existing sparse convolutions for pillars, we implemented SPS-Conv and FS-Conv into the pillar-based object detection backbone. As illustrated in Fig.~\ref{fig:fig_pillar-obj-det}(c), these methods aimed to balance computational savings and model accuracy through adaptive dilation but still failed to provide sufficient \textit{spatial information flow} (SIF)~\cite{chen2022focalconv} that constructs spatial expansion of features of important pillars to supply necessary cues for object detection. Both SubM-Conv and SPS-Conv do not increase receptive fields within a stage, while FS-Conv only allows one-time dilation towards deformable directions. This coarse dilation severely limits SIF, impeding the connection of sparse pillars inside the bounding box unless they proceed through downsampling layers in subsequent stages.

The main challenge lies in augmenting SIF by expanding the foreground pillar group, triggered by dilation and active pillars generated by point clouds. Previous convolution operations have not achieved this, either Dense-Conv facilitating the growth of foreground pillars without discerning background, or SubM-Conv suppressing both background and foreground pillar dilation together. SPS-Conv and FS-Conv enhance only the \textit{coarse-grained} SIF, resulting in insufficient foreground information. Given that such SIF discrepancies, specifically the insufficient dilation of the foreground, hinder appropriate feature extraction for object detection, we propose a novel convolution method that selectively and effectively increases the receptive field for important pillars. As shown in Fig.~\ref{fig:fig_pillar-obj-det}(c), our proposed methods promote the fine-grained dilation of important pillars at every sparse convolution, achieving extreme sparsity while preserving model accuracy.

%% file: Sections/method.tex
\section{Method}
\label{section:method}

To tackle the problem of granting SIF while preserving pillar sparsity, we introduce novel operations: selectively dilated convolution (SD-Conv).
 %Depending on whether we are dilating or pruning, we can appropriately select $g_p$ and $M(\cdot)$ to differentiate between foreground and background pillars. 

\begin{figure}[t]
\centering
\includegraphics[width=\linewidth]{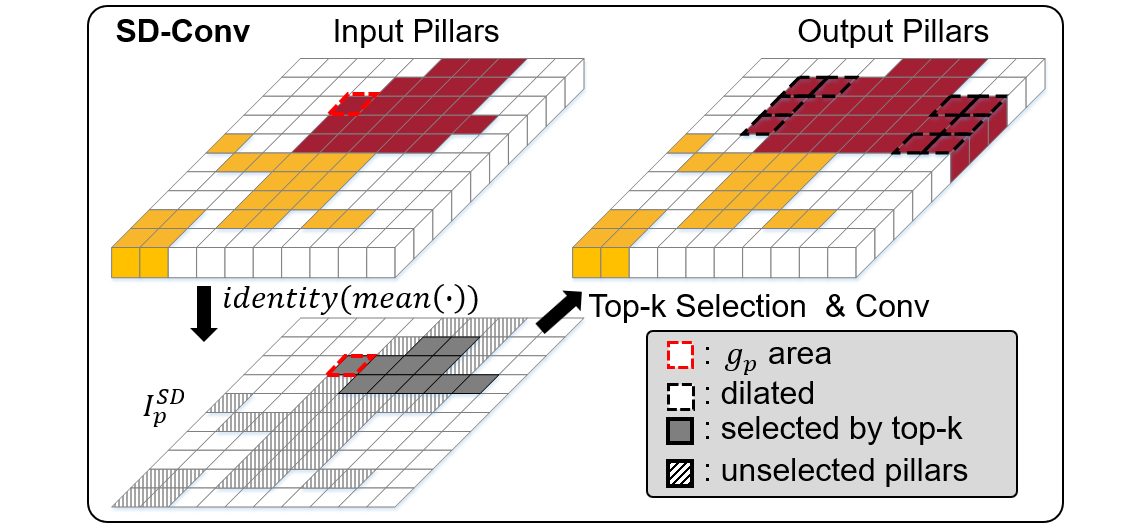}
% \caption{Overview of (a) Selective Dilated Convolution (SD-Conv), and (b) Submanifold Pillar Pruning (SPP).}
\caption{An overview of Selective Dilated Convolution.}
\label{fig_method}
\end{figure}

\begin{table}[t]
\caption{Performance comparison for $t\%$ in top-k selection (BEV Hard mAP of PointPillars on KITTI \textit{val} set.}
\begin{center}
\resizebox{1 \linewidth}{!}{
\renewcommand{\arraystretch}{1.2}
\begin{tabular}{@{\hspace{1pt}}c@{\hspace{1pt}}|@{}c@{\hspace{1pt}}ccccc@{\hspace{1pt}}}
\Xhline{2\arrayrulewidth}
\textit{t}  & \multicolumn{1}{l}{baseline}     & 5    & 4     & 3     & 2     & 1     \\ \hline
mAP $\uparrow$ & 85.63                  & \textbf{85.80} & 85.75 & 85.66 & 85.61 & 83.38 \\ \hline
FLOPs(G) $\downarrow$    & 63.14                  & 5.23 & 4.58  & 4.10  & 3.48  & \textbf{3.06}  \\ \Xhline{2\arrayrulewidth}
\end{tabular}}
\end{center}
\label{tab:t_sweep}
\end{table}

\subsection{Selectively Dilated Convolution}
\label{SD-conv}
Addressing the issue of discontinued SIF, we introduce a novel module, \textit{selectively dilated convolution} (SD-Conv). The foundation of this method is the measurement of pillar importance ($I_p$), defined as follows:
\begin{equation}
\label{eq:PillarImportance}
{I_{p}} = {{G}_{i\in g_p}(M(f_i))},
\end{equation} 
where $g_p$ denotes a group of pillars in the vicinity of a pillar $p$, while $M(\cdot)$ signifies an importance measure applied to a feature vector $f_i\in \mathbb{R}^C$. An overview of these proposed methods is depicted in Fig.~\ref{fig_method}.
% To address the issue of discontinued information flow, we propose a novel module called selectively dilated convolution (SD-Conv) which performs a regular convolution with dilation selectively on the important pillars measured by $I_p$ with the setting of $G_p=\{p\}$ and $M()=mean()$. The intuition is that each pillar should be capable of dilating its feature to its neighbor as long as the feature has enough strength. The important pillars are chosen simply by the top-k method: pillars with top $t\%$ importance are selected for dilation. SubM-Conv is performed for the remaining unimportant pillars. The selection ratio $t$ is kept small to prevent unnecessary dilation and maintain sparsity; we confirm from an ablation study that $t=2\%$ is sufficient for PointPillars~\cite{lang2019pointpillars} on KITTI~\cite{geiger2012kitti}, while $t=4\%$ is preferred for the other cases.
This module performs a regular convolution selectively dilating important pillars, determined by $I_p$, with $g_p={p}$,  $G(\cdot)=identity(\cdot)$ and $M(\cdot)=mean(|\cdot|)$. The intuition is to allow each pillar to expand its feature to its neighbor, provided the feature is strong enough. The ablation study for justification of the proposed importance measure is more discussed in Sec.~\ref{subsec:ablations}. 

For an efficient on-the-fly decision of important pillars, we propose a \textit{dilation threshold} learned throughout training. The top-k method is employed during this training to identify the important pillars, choosing pillars with the top $t$\% importance for dilation, and performing SubM-Conv for the remaining less significant pillars. To prevent unnecessary dilation and preserve sparsity, the selection ratio $t$ is kept small. Through an ablation study in Table~\ref{tab:t_sweep}, we confirmed that $t=2$ is sufficient for PointPillars~\cite{lang2019pointpillars} on KITTI~\cite{geiger2012kitti}, while $t=4$ is preferred in other cases. After training, an importance threshold satisfying the selection ratio becomes the dilation threshold for efficient inference.

\begin{figure}[t]
\centering
\includegraphics[width=\linewidth]{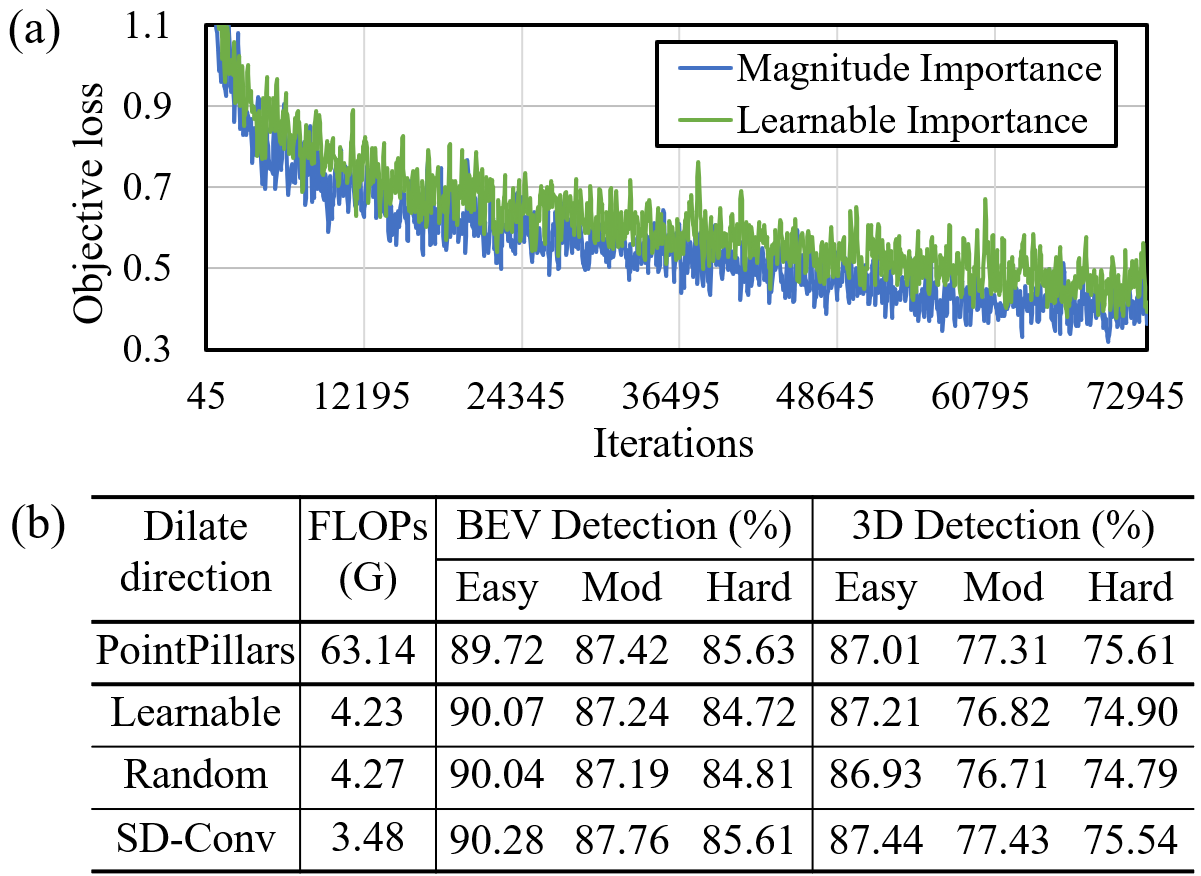}
\caption{(a) Training curve for SparsePointPillars with SD-Conv employing magnitude-based and trainable-based importance in high sparsity. (b) Performance comparison of the car detection task in SparsePointPillars using different methods to determine the dilation directions of SD-Conv applied to the KITTI dataset.}
\label{fig_trainable}
\end{figure}
\begin{figure}[t]
\centering
\includegraphics[width=\linewidth]{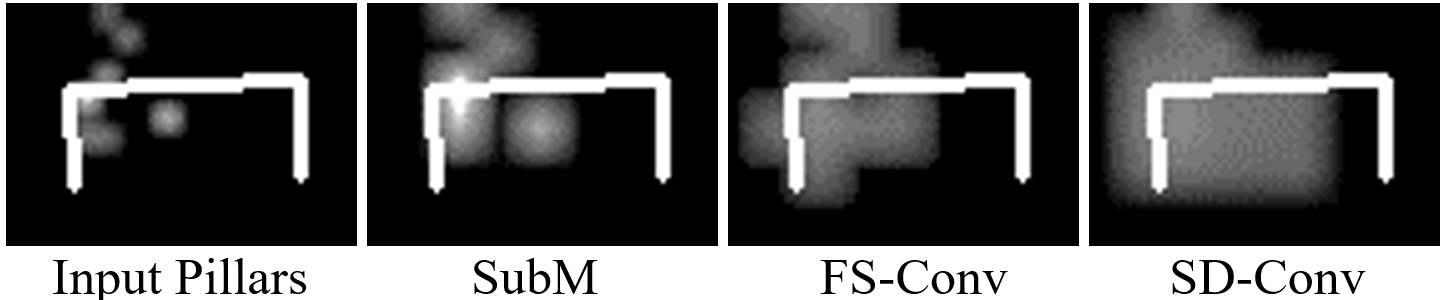}
% \caption{The Feature representation for single object \textit{car}, based on the type of convolution module. Input Pillars is the initial input of backbone network, SubM-Conv, FS-Conv, and SD-Conv are the output of the last layer in stage-2. The white rectangles denote the boundaries of GT-boxes boundary.}
\caption{Feature representation of a single \textit{car} object based on sparse convolution type. ``Input Pillars'' represents the initial input of the backbone network, while the images corresponding to SubM-Conv, FS-Conv, and SD-Conv are the outputs of the last layer in Stage 2. The white rectangles indicate the boundaries of GT-boxes.}
% \caption{This represents the feature representation of a single \textit{car} object, categorized by the convolution module type. The Input Pillars image denotes the initial input for the backbone network, while the images labeled SubM-Conv, FS-Conv, and SD-Conv correspond to the outputs of the final layer in stage-2. The white rectangles indicate GT-box boundaries.}
\label{fig_receptive_effect}
\end{figure}
\begin{figure*}[t]
\centering
\includegraphics[width=\linewidth]{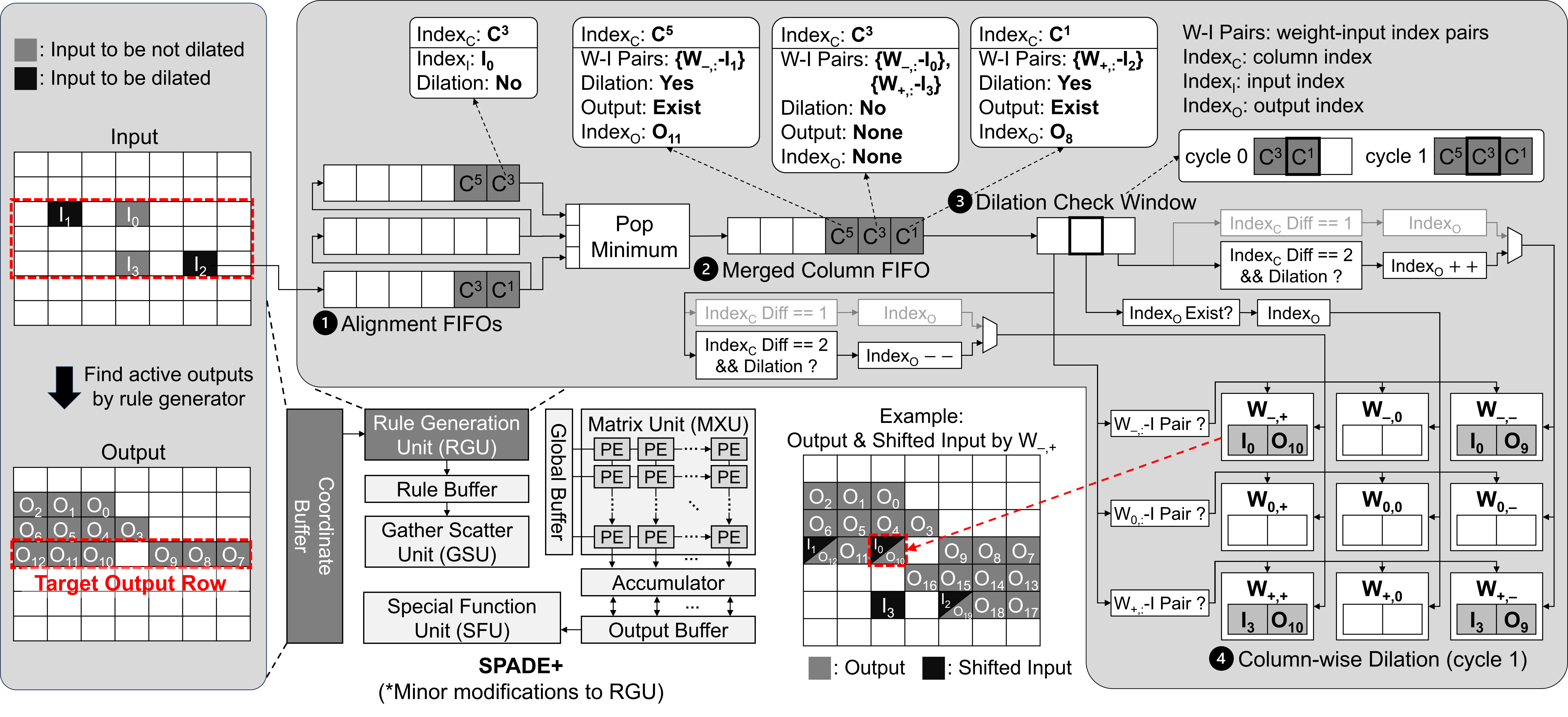}
\caption{An overview of SPADE+ (Minor modification for SD-Conv).}
\label{fig_hw}

\end{figure*}

\subsection{Convergent Selective Dilation}
\label{SD-conv2}
SD-Conv employs two intuitive design aspects: a magnitude-based importance measure and dilation in all directions. These choices contrast with previous studies that advocated for more flexible dilation using learnable parameters. %For instance, \cite{chen2022focalconv} proposed selective dilation with an importance map and dilation directions computed from learnable parameters.
However, our findings showed that this parameterized approach is not suitable for extreme pillar sparsity. 

Fig.~\ref{fig_trainable}(a) presents the training curve of SD-Conv on KITTI for PointPillars with 14\% sparsity, comparing importance measured by either pillar magnitude or learnable parameters. The graph shows that the loss associated with the learnable parameters is consistently higher than that of magnitude-based importance, indicating optimization challenges under significant sparsity. Furthermore, Fig.~\ref{fig_trainable}(b) contrasts 3D object detection accuracy and FLOPs for various dilation direction choices. Surprisingly, random direction dilation outperforms the learned dilation direction in both FLOPs and mAP, while our all-direction dilation yields the best performance.  Consequently, we choose to measure importance based on magnitude and implement dilation in all directions, eliminating additional overhead associated with learnable parameters and dilation choices. As a result, SD-Conv demonstrates higher performance compared to both random and learnable-based dilation directions, with the added benefit of lower FLOPs.

\subsection{Fine-Grain Spatial Information Flow}
\label{SD-conv2}
Fig.~\ref{fig_receptive_effect} contrasts SIF of various sparse convolution methods by observing output features from Stage 2 of PointPillars on KITTI. Specifically, we examine a focused region corresponding to a car (indicated by the white ground truth bounding box). Note that the input pillars are sparsely distributed and disconnected along the bounding box. The SubM-Conv output feature illustrates a discontinuous SIF, failing to provide sufficient features for object detection. FS-Conv, on the other hand, offers a more connected, albeit deformed, SIF that doesn't adequately cover the bounding box. Conversely, our proposed SD-Conv extends the SIF to fill most features within the bounding box, demonstrating that selective dilation within a stage by employing SD-Conv enhances feature provision for object detection.

\section{SPADE+}
\label{SPADE+_hw}
We adapted SPADE~\cite{lee2024spade} to support not only Sparse Convolution, PS-Conv and SubM-Conv, but also SD-Conv, creating new one called SPADE+. SPADE effectively handles operations only for non-zero points but considers dilation for all active points. In contrast, SPADE+ supports the generation of mapping for SD-Conv, where only important active pillars are dilated, through four stages: Alignment (\circled{1}), Row Merge (\circled{2}), Dilation Check Window (\circled{3}), and Column-wise Dilation (\circled{4}) by rule generation unit (RGU). While the mapping operation in the original SPADE only consists of Alignment, Row Merge, and Column-wise Dilation, SPADE+ adds a Dilation Check Window (\circled{3}) between Row Merge and Column-wise Dilation. Additionally, information regarding dilation status is added to the FIFOs of each stage in SPADE+, facilitating the generation of mapping information based on dilation status.

% We've adjusted only the RGU segment of SPADE to support SD-Conv. 
% Unlike SPADE, where all columns undergo dilation or remain undilated, in SPADE+, only important input pillars dilate to generate output. 
In SPADE+, only important input pillars dilate to generate output, so output index calculation depends on whether the merged column index dilates, determined by neighboring columns within Dilation Check Window (\circled{3}). For dilated merged columns, the SPADE+ behaves like SPADE. For non-dilated ones, rule generation depends on nearby columns' information. 
As illustrated in the example of Fig.~\ref{fig_hw}, when generating the mapping for the 4-th target output row, the Merge Row results in a total of three merged columns. Subsequently, in cycle 0, column-wise dilation is applied to the first merged column, followed by column-wise dilation for the second merged column in cycle 1.
% For example, as shown in Fig.~\ref{fig_hw}, merged column of $I_0$ and $I_3$ 
% In cycle 1, merged column ($C^3$) of $I_0$ and $I_3$ does not produce any output, so no mapping is created for $W_{:,0}$. 
In cycle 1, since $I_0$ and $I_3$ are not dilated and there are no adjacent columns with merged columns, no output is generated for the merged column ($C^3$), resulting in the mapping for $W_{:,0}$ not being created. However, for $W_{:,-}$ and $W_{:,+}$, adjacent columns are expanded to generate mapping information. 
As a result, it can be observed that $I_0$ is shifted by $W_{-,+}$ to be computed with $O_{10}$, indicating that when the positions of all outputs are known, $I_0$ is indeed shifted to the $O_{10}$ by $W_{-,+}$.
With minor tweaks to the rule generation unit, the SPADE+ operates akin to SPADE, facilitating SD-Conv operations.

SPADE+ retains the rest of SPADE's components unchanged, except for the RGU, which is modified to generate mapping information for SD-Conv. To streamline mapping in SPADE~\cite{lee2024spade}, it simultaneously identifies output positions from convolution operations and generates relevant mapping details using a RGU. 
% The mapping operation involves three steps: alignment, row merge, and column-wise dilation. The latter two steps consider both row-wise and column-wise dilations separately to determine the relationship between sparse input and output for convolution kernels. 
Additionally, a gather-scatter unit manages input, weight, and output to minimize data movement during convolutions, enhancing performance based on sparsity. Furthermore, in SPADE, PS-Conv was proposed to minimize computational overhead without sacrificing performance. However, as evident from the Table~\ref{tab:pp} results, PS-Conv fails to address the discontinued SIF issue properly, resulting in higher computational complexity compared to SD-Conv.

Since SPADE+ only modifies the RGU portion to support SD-Conv, the area overhead of SPADE+ is only about 1\% compared with original SPADE~\cite{lee2024spade} as shown in the Table~\ref{hw_comp}. When comparing the effective TOP/W of SPADE for PS-Conv and SPADE+ for SD-Conv, the latter utilizing SD-Conv shows a $3.5\times$ improvement thanks to the small hardware overhead and computation efficiency of SD-Conv.

% with just 1\% area overhead as shown in Table~\ref{hw_comp}. 

\begin{table}[t]
\caption{Performance comparison of PointPillars among various types of sparse convolution (KITTI \textit{val} set).}
\begin{center}
\resizebox{\linewidth}{!}{
\renewcommand{\arraystretch}{1.1}
\begin{tabular}{@{\hspace{3pt}}c@{\hspace{3pt}}|l|@{\hspace{3pt}}c@{\hspace{3pt}}|ccc}
\Xhline{2\arrayrulewidth} 
\multirow{2}{*}{Method}                                                        &  \multirow{2}{*}{Conv type}     & \multirow{2}{*}{\begin{tabular}[c]{@{}c@{}}FLOPs\\ (G)$\downarrow$\end{tabular}} & \multicolumn{3}{c}{3D Detection (\%)} $\uparrow$  \\ \cline{4-6} 
                                                                               &                                &                                                                                                                                                & Easy   & Mod    & Hard    \\ \Xhline{2\arrayrulewidth} 
PointPillars                                                                   & Dense                                                                                       & 63.14                                                                 & 87.01  & 77.31 & 75.61 \\ \hline
\multirow{5}{*}{\begin{tabular}[c]{@{}c@{}}Sparse\\ PointPillars\end{tabular}} & SubM-Conv                                                                                         & \textbf{2.65}                                                                  & 87.35  & 74.91 & 72.45 \\ \cline{2-6}
 & PS-Conv                                                                                         & 16.73                                                                  & 86.10  & 76.94 & 74.15 \\ \cline{2-6}
                                                                               & \multirow{2}{*}{FS-Conv}                                                                           & 5.99                                                                  & 87.09  & 76.82 & 75.09 \\
                                                                               &                                                                                                    & 4.60                                                                   & 87.16  & 76.85 & 72.64 \\ \cline{2-6} 
                                                                               & SD-Conv                                                                                            & 3.48                                                                  & \textbf{87.44}  & \textbf{77.43} & \textbf{75.54} \\ \Xhline{2\arrayrulewidth} 
\end{tabular}}
\end{center}
\label{tab:pp}
\end{table}

\begin{table}[]
\caption{Hardware comparison between SPADE and SPADE+.}
\label{hw_comp}
\begin{center}
\resizebox{1\linewidth}{!}{
\renewcommand{\arraystretch}{1.2}
\begin{tabular}{c|cc}
\Xhline{2\arrayrulewidth} 
\multicolumn{1}{l|}{}                                         & \multicolumn{1}{c|}{SPADE}                                                                    & SPADE+                                                                                                  \\ \Xhline{2\arrayrulewidth} 
Cores                                                          & \multicolumn{2}{c}{$64\times64=4096$}                                                                                                                                                                         \\ \hline
SRAM (KB)                                                      & \multicolumn{2}{c}{654}                                                                                                                                                                                \\ \hline
Area ($mm^2$)                                    & \multicolumn{1}{c|}{11.6}                                                                     & 11.73                                                                                                   \\ \hline
\begin{tabular}[c]{@{}c@{}}Effective\\ TOP/W\end{tabular}                            & \multicolumn{1}{c|}{6.27}  
 & 22.13 \\ \hline
\begin{tabular}[c]{@{}c@{}}Supported \\ Conv Type\end{tabular} & \multicolumn{1}{l|}{\begin{tabular}[c]{@{}l@{}}Sparse Conv, PS-Conv,\\ SubM-Conv\end{tabular}} & \multicolumn{1}{l}{\begin{tabular}[c]{@{}l@{}}Sparse Conv, PS-Conv,\\ SubM-Conv, \textbf{SD-Conv}\end{tabular}} \\ \Xhline{2\arrayrulewidth} 
\end{tabular}}
\end{center}
\vspace{-0.3 cm}
\end{table}

%% file: Sections/experiments.tex
\section{Experiments}
%In this section, we evaluate the SD-Conv's performance
\subsection{Experimental Setting}

We employed three state-of-the-art pillar-based 3D object detection networks, PointPillars (PP)~\cite{lang2019pointpillars}, CenterPoint (CP)~\cite{yin2021center}, and PillarNet (PN)~\cite{shi2022pillarnet}, for evaluation of the proposed method on KITTI~\cite{geiger2012kitti} (for PP) or nuScenes~\cite{caesar2020nuscenes} (for CP and PN) benchmarks. We followed the baseline settings of SparsePointPillars~\cite{vedder2021sparse} to replace the existing convolution operations (Conv2D and SubM-Conv) of PP, CP, and PN with SD-Conv. All the experimental settings are implemented with PyTorch-based frameworks, including OpenPCDet\footnote{https://github.com/open-mmlab/OpenPCDet} for PP and the popular CenterPoint\footnote{https://github.com/tianweiy/CenterPoint} code-base for PN and CP, and run on the NVIDIA A100 GPU.

\subsection{Main Results}

\textbf{PointPillars (PP): }
% We first compare the performance of the proposed SD-Conv with the other sparse convolution operations (SubM-Conv and FS-Conv) on PP. As Table~\ref{tab:pp} illustrates, SubM-Conv dramatically reduces the number of computations (FLOPs) by 95\% at the cost of significant accuracy degradation, especially for the Hard category (-3.16 mAP), indicating that the restricted SIF causes malicious impact on the object detection. We sweep target FLOPs for both FS-Conv and SD-Conv; we found that FS-Conv requires 5.99 GFLOPs to maintain the object detection accuracy, and the FLOPs lower than that results in significant accuracy degradation like SubM-Conv. Whereas SD-Conv can reduce the FLOPs by 94.5\% while maintaining the original accuracy, indicating that the proposed fine-grained selective dilation can be beneficial for constructing necessary SIF to identify objects properly. 
We begin by comparing the performance of our proposed SD-Conv with other sparse convolution operations, SubM-Conv, PS-Conv and FS-Conv, on the PointPillars (PP) with the popular KITTI dataset. As shown in Table~\ref{tab:pp}, SubM-Conv achieves a remarkable 95\% reduction in computational operations (FLOPs). However, this comes at the cost of a significant decrease in accuracy, particularly evident in the Hard category (-3.16 mAP). This outcome underscores the adverse impact of constrained SIF caused by SubM-Conv on object detection. PS-Conv demonstrates a lesser accuracy degradation in the Hard category, with a -1.46 mAP decrease. While PS-Conv alleviates accuracy degradation to some extent, it offers a significantly lesser reduction in computational operations compared to the other methods. This highlights that merely applying pruning while concurrently training does not suffice to efficiently maintain SIF. To improve accuracy while saving computations, we systematically vary the target FLOPs for both FS-Conv and SD-Conv. Notably, FS-Conv demands 5.99 GFLOPs to maintain the original object detection accuracy. Any reduction in FLOPs below this threshold results in a substantial decline in accuracy, similar to what we observed with SubM-Conv. Conversely, SD-Conv achieves a remarkable 94.5\% reduction in FLOPs while preserving the original accuracy. This demonstrates the effectiveness of our proposed fine-grained selective dilation in constructing essential SIF, thereby enabling accurate object identification.

\begin{table*}[t]
\caption{Performance comparison between CenterPoint and its variants on the nuScenes \textit{val} set.}
\begin{center}
\resizebox{0.81\linewidth}{!}{
\renewcommand{\arraystretch}{1.2}
\begin{tabular}{cl|l|c|c|c|ccccc}
\Xhline{2\arrayrulewidth}
\multicolumn{2}{c|}{Model}                                                                             & Conv type  & \begin{tabular}[c]{@{}c@{}}FLOPs\\ (G) $\downarrow$ \end{tabular} & mAP $\uparrow$  & NDS $\uparrow$   & mATE  & mASE  & mAOE  & mAVE  & mAAE $\downarrow$  \\ \Xhline{2\arrayrulewidth}
\multicolumn{2}{c|}{CenterPoint}                                                                    & Dense-Conv                                                                 & 70.01                                              & 50.79 & 60.55 & 31.77 & 25.97 & 35.94 & 34.77 & 19.97 \\ \hline
\multicolumn{2}{c|}{\multirow{4}{*}{\begin{tabular}[c]{@{}c@{}}Sparse \\ CenterPoint\end{tabular}}} & SubM-Conv                                                                & \textbf{18.13}                                               & 47.89 & 58.94 & 32.04 & 26.32 & 36.83 & 34.53 & 20.36 \\ \cline{3-11} 
\multicolumn{2}{c|}{}                                                                                  & PS-Conv                                                             & 27.09                                               & 50.12 & 60.42 & 31.38 & 26.24 & 36.34 & 32.27 & 19.73 \\ \cline{3-11} 
\multicolumn{2}{c|}{}                                                                                  & FS-Conv                                                             & 23.34                                               & 50.30 & 60.41 & 31.55 & 26.10 & 38.14 & 32.67 & 19.95 \\ \cline{3-11}
\multicolumn{2}{c|}{}                                                                                  & SD-Conv                                                             & 19.37                                               & \textbf{50.33} & \textbf{60.84} & \textbf{31.28} & \textbf{25.96} & \textbf{34.26} & \textbf{32.11} & \textbf{19.66} \\ \Xhline{2\arrayrulewidth}
\end{tabular}}
\end{center}
\label{tab:cp}
\end{table*}

% We initially compared the performance of our proposed method, SD-Conv on PP, with sparse convolution operations: SubM-Conv and FS-Conv. As Table~\ref{tab:pp} illustrates, SubM-Conv resulted in a notable performance drop, with a 3.16 mAP decrease in 3D hard tasks at 2.65 GFLOPs. FS-Conv saw a minor accuracy reduction (0.52 mAP) with moderate computation density, but further savings led to a substantial accuracy loss. Conversely, implementing SD-Conv maintained comparable (or even superior) accuracy to the baseline while achieving a significant reduction in computation, up to 18 times. Note that the reason for such a drastic reduction in computational cost is that in the case of PointPillars (Dense-Conv), the deconv layer of PP occupies 1/3 of the total FLOPs. However, since this deconv layer performs very sparse operations, Sparse PointPillars can achieve a significant FLOPs reduction. The performance improvement demonstrates the potential of our approach to enhance model performance efficiently and effectively.

%Conversely, implementing SD-Conv maintained comparable (or superior) accuracy to the baseline while significantly reducing computation. Applying SPP to SD-Conv achieved computation savings of over 10 times, maintaining equal accuracy. Further reducing computation density below 9\% resulted in a less than 1.00 mAP decrease in 3D object detection, which could be effectively compensated for by fine-tuning the model post-pruning. The performance improvement demonstrates the potential of our approach to enhance model performance efficiently and effectively.
\begin{table}[t]
\caption{Performance comparison between PillarNet and its variants on the nuScenes \textit{val} set.}
\begin{center}
\resizebox{1\linewidth}{!}{
\renewcommand{\arraystretch}{1.3}
\begin{tabular}{@{\hspace{2pt}}c@{\hspace{2pt}}|@{\hspace{2pt}}c@{\hspace{2pt}}|@{\hspace{2pt}}c@{\hspace{2pt}}|c|c|@{\hspace{4pt}}c@{\hspace{4pt}}|@{\hspace{4pt}}c@{\hspace{4pt}}}
\Xhline{2\arrayrulewidth}
\multirow{2}{*}{Model}                                                       & Backbone & Neck & Head & \multirow{2}{*}{\begin{tabular}[c]{@{}c@{}}FLOPs\\ (G) \end{tabular}} & \multirow{2}{*}{mAP} & \multirow{2}{*}{NDS} \\ \cline{2-4}     &       \multicolumn{3}{@{\hspace{2pt}}c|}{Conv type} &                     &                      &              \\ \Xhline{2\arrayrulewidth}
PilllarNet                                                                                  & SubM & Desne & Desne                                                                      & 276.26                                                              & 59.58                & 66.95        \\ \hline
\multirow{2}{*}{\begin{tabular}[c]{@{}c@{}}Sparse\\ PillarNet\end{tabular}}               & SubM    & SubM     &Desne                                                                     & \textbf{151.96}                                                               & 57.92                & 66.33                         \\ \cline{2-7} 
                                                                               & SubM & SD-Conv               & Desne                                                                     & 162.29                                                               & \textbf{59.45}                & \textbf{67.40}                            \\  \Xhline{2\arrayrulewidth}
\end{tabular}}
\end{center}
\label{tab:pn}
\vspace{-0.5cm}
\end{table}

\textbf{CenterPoint (CP): }
% We further evaluate the performance of SD-Conv on another popular pillar-based 3D object detection, CenterPoint (CP). Table~\ref{tab:cp} displays total FLOPs, and the 3D object detection accuracies in seven categories following the convention of nuScenes \textit{val} set. SubM-Conv reduces 74.1\% of computations at the cost of noticeable accuracy degradation for all the performance categories. In comparison, FS-Conv maintains the original accuracy while saving only up to 66.7\% of computations. Conversely, SD-Conv can safely reduce 72.3\% of computation while achieving accuracy superior to FS-Conv, indicating superior trade-off of SD-Conv between FLOPs and accuracy thanks to its fine-grain SIF construction.
We evaluate SD-Conv using CenterPoint (CP), another popular pillar-based 3D object detection framework. The results in Table~\ref{tab:cp} show the FLOPs and 3D object detection accuracies in two categories, and errors in five categories, following the nuScenes \textit{val} set convention. In the case of SubM-Conv, we observe a substantial 74.1\% reduction in computations, but this comes at the cost of noticeable accuracy degradation across all categories. PS-Conv reduces computational load by 61.28\%, which is smaller than the reduction achieved by SubM-Conv, but it demonstrates improved accuracy. In contrast, FS-Conv maintains original accuracy while achieving only up to a 66.7\% reduction in computational load. SD-Conv, on the other hand, safely reduces computation by 72.3\% while surpassing FS-Conv in terms of accuracy. This highlights the advantageous trade-off provided by SD-Conv between FLOPs and accuracy, primarily due to its fine-grained SIF construction.
 
% We evaluated the performance of SD-Conv on CP and PN to assess the broad applicability of our methods. Table~\ref{tab:cp} displays total FLOPs, and CP inference accuracy on the nuScenes \textit{val} set. SubM-Conv reduced computation to 27\% but at a significant accuracy loss (1.90 mAP and 1.61 NDS). FS-Conv demonstrated limited computation savings while maintaining baseline performance. In contrast, SD-Conv achieved similar computation savings to SubM-Conv without an accuracy trade-off.

% \textbf{PillarNet (PN): }
% We further evaluate the benefit of SD-Conv on the state-of-the-art pillar-based object detection, PillarNet (PN). PN consists of not only the backbone, neck, and head where backbone includes multiple layers of SubM-Conv to enhance feature extraction capability. Since the most portion of computation is taken by the neck, we focus on sparsifying it with either SubM-Conv or SD-Conv to reduce overall FLOPs while maintaining accuracy. In Table~\ref{tab:pn}, we report the representative accuracy metrics, mAP and NDS with the nuScene \textit{val} dataset, but the overall trend is the same. Replacing the dense convolution in the neck of PN with the SubM-Conv leads to significant 45.0\% of computational savings while the mAP is noticeably degraded. However, converting SubM-Conv into SD-Conv results in full recovery of the accuracy (even superior NDS) with marginal increase of FLOPs. These results highlight the broad applicability of SD-Conv in maintaining or improving performance while reducing computational cost.
\textbf{PillarNet (PN): }
We further assessed SD-Conv's benefits in the context of PillarNet (PN), a cutting-edge pillar-based object detection method, where its backbone incorporates multiple layers of SubM-Conv to enhance feature extraction. Given that a significant portion of computation resides in its neck, our focus was on sparsifying it using both SubM-Conv and SD-Conv to reduce overall FLOPs while preserving accuracy. Table~\ref{tab:pn} reports key accuracy metrics, including mAP and NDS, using the nuScene \textit{val} dataset, with consistent overall trends. Replacing the dense convolution in PN's neck with SubM-Conv results in substantial computational savings of 45.0\% but noticeable mAP degradation. Conversely, transitioning from SubM-Conv to SD-Conv fully restores accuracy (its NDS even surpasses the baseline) with only a marginal increase in FLOPs compared to . These findings underscore SD-Conv's versatile applicability in maintaining or enhancing performance while simultaneously reducing computational overhead.

\begin{table}[t]
\caption{Ablation study of importance metric of SD-Conv (Sparse PointPillars with KITTI \textit{val} set).} 
\begin{center}
\resizebox{\linewidth}{!}{
\renewcommand{\arraystretch}{1.2}
\begin{tabular}{@{\hspace{2pt}}l@{\hspace{2pt}}|@{\hspace{3pt}}l@{\hspace{3pt}}|@{\hspace{2pt}}l@{\hspace{2pt}}|@{\hspace{2pt}}c@{\hspace{2pt}}|ccc}
\Xhline{2\arrayrulewidth}
                   \multirow{2}{*}{$M(\cdot)$}    & \multirow{2}{*}{$G(\cdot)$}         & \multirow{2}{*}{$g_p$}       & \multirow{2}{*}{\begin{tabular}[c]{@{}c@{}}FLOPs\\ (G)\end{tabular}}  & \multicolumn{3}{c}{3D Detection (\%)} \\ \cline{5-7} 
                                           &                       &                            &                                                                                                  & Easy   & Mod   & Hard  \\\Xhline{2\arrayrulewidth}
   Mean                  & \textit{identity}                   & $p$                        & 3.48                                                                  &\textbf{87.44}  & \textbf{77.43} & \textbf{75.54} \\
                                            Max                   & \textit{identity}                  & $p$                        & \textbf{3.07}                                                                   & 87.62  & 76.93 & 72.67 \\
                                           Mean                  & Avg-Pool                   & SubM($p$)                  & 3.34                                                                   & 86.99  & 76.49 & 72.84 \\
                                            Mean                  & Max-Pool                   & SubM($p$)                  & 3.43                                                                   & 87.14  & 77.12 & 74.73 \\ \Xhline{2\arrayrulewidth}

\end{tabular}}
\end{center}
\label{tab:imp}
\vspace{-0.5cm}
 \end{table}

\subsection{Ablation Study}
\label{subsec:ablations}

In this section, we validate several design choices for our pillar-based 3D object detection: 1) the metric for measuring importance of SD-Conv, 2) types of sparse convolution for down-sampling, and 3) methods for enhancing SIF.

% \begin{wrapfigure}{r}{0.45\textwidth}  % 'l' for left alignment, '0.5\textwidth' for half of the text width
% \vspace{-0.7 cm}
%   \includegraphics[width=\linewidth]{Figure/spar_acc.png}
  
%   \caption{Sparsity vs Performance}
% \label{spar_acc}
% \vspace{-0.4 cm}
% \end{wrapfigure}

% \textbf{Sparsity vs. Performance: } As shown in Fig.~\ref{fig_spar_acc2}, We evaluate the performance of employing SD-Conv in PP~\cite{lang2019pointpillars} to enhance sparsity, compared with the results of FS-Conv. Despite maintaining performance up to 3.48 GFLOPs with SD-Conv, FS-Conv shows significant performance degradation at 4.60 GFLOPs, demonstrating the challenge of increasing sparsity. It also underscores the importance of the SIF through dilation across all layers.

\textbf{Importance Metric: } Table~\ref{tab:imp} presents findings regarding various metrics related to $I_p$, as discussed in Sec.~\ref{section:method}. We explore different metrics within a magnitude-based approach for $I_p$. Regarding the selection of important pillars, the mean across the channel consistently demonstrates the best performance. Conversely, the max metric tends to excessively emphasize outliers. Average pooling (Avg-Pool) faces challenges in making accurate assessments when neighboring values of a particular pillar were zero, diminishing its relevance. Meanwhile, max pooling (Max-Pool) underperforms as it places excessive emphasis on high-magnitude pillars.

\textbf{Types of Down-Sample Sparse Convolution: }
As discussed in Sec.~\ref{subsec:pillar-based-3d-object-detection}, down-sample sparse convolution at the beginning of stages increases the receptive field, affecting the pillar-based object detection's overall sparsity. Two down-sampling methods exist: sparse convolution with a 2x2 kernel used by SparsePointPillars~\cite{vedder2021sparse} and spatial pruned regular sparse convolution (SPRS-Conv~\cite{liu2022spatial}). The former increases sparsity by shrinking the dilation window from 3x3 to 2x2. Meanwhile, SPRS-Conv dilates only vital features, risking pruning essential elements due to its pre-feature-importance stride mask. Table~\ref{tab:strided_conv} shows that SPRS-Conv results in lower accuracy than the case with a 2x2 kernel window, thus we employed the 2x2 kernel approach in our work.

\begin{table}[t]
\caption{Comparison of down-sample sparse conv: 2x2 kernel vs. SPRS-Conv (Sparse PointPillars on KITTI \textit{val} set).}
\begin{center}
\resizebox{1\linewidth}{!}{
\renewcommand{\arraystretch}{1.2}
\begin{tabular}{c|l|c|ccc}
\Xhline{2\arrayrulewidth}
\multirow{2}{*}{Model}                                                                    &  \multirow{2}{*}{\begin{tabular}[c]{@{}c@{}}Down-Sample\\ Conv Type\end{tabular}} & \multirow{2}{*}{\begin{tabular}[c]{@{}c@{}}FLOPs\\ (G)\end{tabular}} & \multicolumn{3}{c}{3D Detection (\%)} \\ \cline{4-6} 
                                                                                          &                                                                              &                                                                         & Easy   & Mod   & Hard  \\ \Xhline{2\arrayrulewidth}
\multirow{2}{*}{\begin{tabular}[c]{@{}c@{}} PointPillars \\+ SD-Conv\end{tabular}} & 2x2 kernel~\cite{vedder2021sparse}                                                              & \textbf{3.48}                                                                   & \textbf{87.44}  & \textbf{77.43} & \textbf{75.54} \\ \cline{2-6} 
                                                                                          & SPRS-Conv~\cite{liu2022spatial}                                                                    & 3.57                                                                   & 87.39  & 76.92 & 74.16 \\ \Xhline{2\arrayrulewidth}
\end{tabular}}
\end{center}
\label{tab:strided_conv}
\vspace{-0.5cm}
\end{table}

\begin{table}[t!]

\caption{Comparison of additional down-sampling (ADS~\cite{chen2023voxelnext}) and SD-Conv (PointPillars on KITTI \textit{val} set).}
\begin{center}
\resizebox{0.85\linewidth}{!}{
\renewcommand{\arraystretch}{1.1}
\begin{tabular}{c|c|ccc}
\Xhline{2\arrayrulewidth}
\multirow{2}{*}{Method}         & \multirow{2}{*}{\begin{tabular}[c]{@{}c@{}}FLOPs\\ (G)\end{tabular}}  & \multicolumn{3}{c}{3D Detection (\%)} \\ \cline{3-5} 
                                &                                                                             &                                                                          Easy   & Mod    & Hard  \\ \Xhline{2\arrayrulewidth}
Dense                                                                                                  & 63.14                                                                  & 87.01  & 77.31  & 75.61 \\ \hline
SubM                                                                                              & \textbf{2.65}                                                                  & 87.35  & 74.91  & 72.45 \\ \hline
SubM + ADS                                                                      & 4.10                                                                  & 87.06  & 76.23  & 72.37 \\ \hline
SD-Conv                                                                                              & 3.48                                                                   & \textbf{87.44}  & \textbf{77.43}  & \textbf{75.54} \\ \Xhline{2\arrayrulewidth}
\end{tabular}}
\end{center}
\label{tab:add-down}
\vspace{-0.5cm}
\end{table}

% \begin{table}[t]
% \caption{Performance comparison for the $t\%$ in top-k selection on KITTI \textit{val} set with PointPillars.}
% \begin{center}
% \resizebox{1 \linewidth}{!}{
% \renewcommand{\arraystretch}{1}
% \begin{tabular}{c|cccccc}
% \Xhline{2\arrayrulewidth}
% \textit{t}  & \multicolumn{1}{l}{baseline}     & 5    & 4     & 3     & 2     & 1     \\ \hline
% BEV Hard mAP & 85.63                  & \textbf{85.80} & 85.75 & 85.66 & 85.61 & 83.38 \\ \hline
% FLOPs(G)    & 63.14                  & 10.08 & 9.43  & 8.95  & 8.33  & \textbf{7.91}  \\ \Xhline{2\arrayrulewidth}
% \end{tabular}}
% \end{center}
% \label{tab:t_sweep}
% \vspace{-0.5 cm}
% \end{table}

\textbf{SD-Conv vs. Additional Down-Sampling for SIF: }
Computation savings from SubM-Conv restrict the SIF, compromising object detection accuracy. A recent method, VoxelNext~\cite{chen2023voxelnext}, introduces an additional down-sampling (ADS) to enhance SIF in SubM-Conv. While ADS can boost SIF via a larger receptive field, it also increases the computational load. To compare SD-Conv with ADS, we adjusted Sparse PointPillars to include an extra down-sampling and SubM-Conv for ADS. Table~\ref{tab:add-down} shows SubM+ADS's mixed impact on 3D object detection accuracy (a +1.32 increase on Moderate but drops of -0.29 on Easy and -0.08 on Hard) at the cost of additional 1.45G FLOPs of computation. In contrast, SD-Conv matches the dense baseline's accuracy but with fewer computations than SubM+ADS, underscoring SD-Conv's advantages.

\subsection{Hardware Evaluation}
\label{subsec:hardware}

% Exploiting sparsity of point cloud processing for runtime savings in conventional hardware like GPU is challenging due to lack of architectural supports. To overcome this architectural limitation, several promising point cloud-based sparse convolution accelerators have been proposed, which are equipped with dedicated hardware logic for handling sparse data structures~\cite{feng2020mesorasi,lin2021pointacc,feng2022crescent,lee2023PillarAcc}. To evaluate the feasibility and system-level benefits of the proposed SD-Conv, we devise cycle-accurate simulators for the recent and representative point cloud accelerators~\cite{lin2021pointacc, lee2023PillarAcc}. These point cloud accelerators natively supports sparse convolution by generating mapping information between active inputs and active outputs, enabling it to process only non-zero value GEMM operations. 

Harnessing the sparsity of point cloud processing for runtime savings on conventional GPUs is ineffective due to architectural constraints. However, several point cloud-based sparse convolution accelerators, featuring dedicated logic for sparse data structures, have been introduced~\cite{feng2022crescent,feng2020mesorasi,lee2024spade,lin2021pointacc}. These accelerators support sparse convolution by mapping active inputs to outputs, focusing only on non-zero value GEMM operations. To assess the proposed SD-Conv's feasibility and benefits, we created cycle-accurate simulators for recent point cloud accelerators~\cite{lee2024spade,lin2021pointacc}.

% Finding mapping information involves calculating input-weight-output index pairs, determining which output is generated when each input operates with a specific kernel index. In other words, having mapping information allows us to store the output generated by multiplying each weight with each active point at the appropriate location, effectively eliminating unnecessary calculations and enabling efficient utilization in sparse convolution. On the GPU, hash tables are typically used to generate mapping information for output indices. However, this can lead to collisions during hash table queries and inserts, and due to limitations in parallel collision resolution, it can increase mapping overhead. To overcome this challenge, PointAcc~\cite{lin2021pointacc} uses a merge sorting approach instead of hash tables, while PillarAcc~\cite{lee2023PillarAcc} employs a pipelined structure based on sorted inputs within a limited area to reduce mapping overhead. Additionally, PillarAcc operates without using the cache employed by PointAcc, leveraging mapping information to process data deterministically and using scratch pads to minimize memory overhead resulting from cache misses. 

More specifically, mapping information calculates input-weight-output index tuples, indicating the output from each input with a specific kernel index. This allows for efficient storage of each weight's product with an active point, optimizing sparse convolution calculations. While GPUs often use hash tables to create this mapping, leading to collisions and increased overhead, PointAcc~\cite{lin2021pointacc} employs merge sorting, and SPADE~\cite{lee2024spade} uses a pipelined strategy with sorted inputs to reduce overhead. Unlike PointAcc, which uses a cache, deterministically processes the mapping details and employs scratch pads to minimize memory overhead. Due to these architectural enhancements, both PointAcc and SPADE achieve sparsity-related speedup, as verified in our cycle-accurate simulators.

\begin{figure}[t]
\centering
\includegraphics[width=1\linewidth]{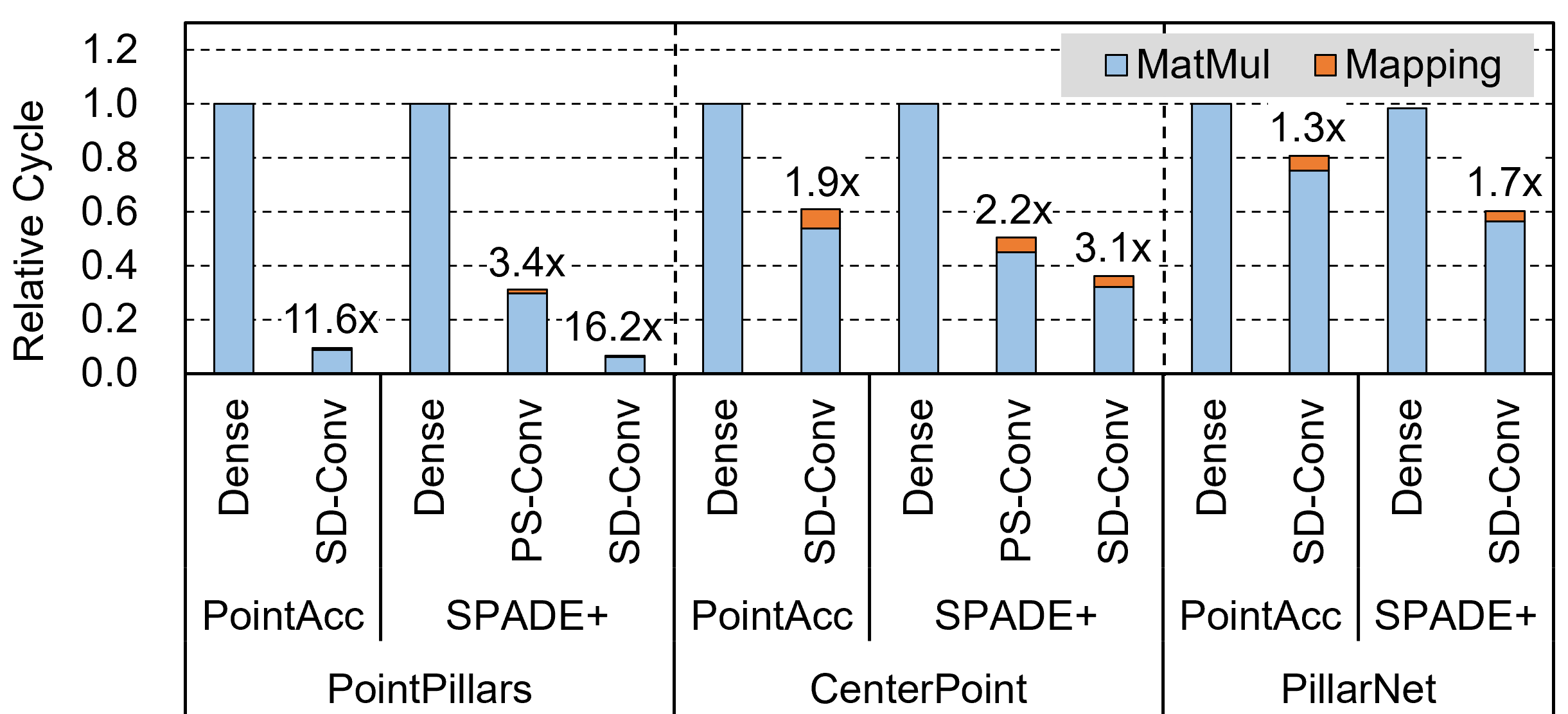}

\caption{Comparison of relative cycle with and without SD-Conv in embedded accelerators (SPADE+, PointAcc), using dense convolution accelerators as the baseline.}

\label{fig_hardware}
\end{figure}

To assess the hardware speedup of SD-Conv, we implement SD-Conv's mapping algorithm into our PointAcc/SPADE simulators and tested it on three models (PP, CP, PN). These simulators also feature a baseline architecture for executing dense convolution in a systolic manner. Fig.~\ref{fig_hardware} shows relative execution cycles compared to the baseline, with SD-Conv configurations as per Tables\ref{tab:pp}-\ref{tab:pn}. For PP, CP, and PN, SPADE+ attains a 16.2$\times$, 3.1$\times$, 1.7$\times$ speedup, nearing the ideal 18.1$\times$, 3.6$\times$, 1.7$\times$ speedup from computational savings, respectively. PointAcc lags behind SPADE+ in all tests due to cache misses but still significantly outperforms the baseline using sparsity.

%% file: Sections/conclusion.tex
\section{Conclusion}

This research demonstrated that pillar-based 3D object detection, an efficient approach in autonomous driving technology, outperforms point-based and voxel-based methods in speed and accuracy, despite the computational redundancy from densifying the intrinsically sparse pillar data. We discovered that the accuracy loss in recent submanifold convolution (SubM-Conv) methods is due to their limited receptive field. To address this, we introduced a selectively dilated convolution (SD-Conv) that enhance accuracy by focusing on key pillars and eliminating non-essential ones. Evaluation across several state-of-the-art models validated that our approach maintains superior sparsity without sacrificing mAP.